\documentclass[letterpaper, 10 pt, conference]{ieeeconf}  
\IEEEoverridecommandlockouts                              
\let\labelindent\relax
\usepackage[english]{babel}
\usepackage[T1]{fontenc}
\usepackage[ruled,vlined]{algorithm2e}
\usepackage{geometry}
\usepackage{amsmath}
\usepackage{amsfonts}
\usepackage{multirow}
\usepackage{enumitem}
\usepackage{graphicx}
\usepackage{subfigure}
\usepackage[utf8]{inputenc} 
\usepackage{hyperref}       
\usepackage{url}            
\usepackage{booktabs}       
\usepackage{nicefrac}       
\usepackage{microtype}      
\usepackage{gensymb}
\usepackage{float}
\usepackage[table,xcdraw]{xcolor}
\usepackage{caption}
\usepackage{color}
\usepackage{amssymb}
\usepackage{pifont}
\usepackage[table,xcdraw]{xcolor}
\usepackage{algpseudocode}
\usepackage{bm}
\usepackage{cite}
\usepackage{stfloats}
\usepackage{bbm}
\usepackage{collcell}

\author{Qiang Wang, \textit{Student Member, IEEE}, Robert McCarthy, David Cordova Bulens, \\Francisco Roldan Sanchez, Kevin McGuinness, Noel E. O’Connor, \\ and Stephen J. Redmond, \textit{Senior Member, IEEE}
\thanks{*This publication has emanated from research conducted with the financial support of China Scholarship Council under grant number CSC202006540003 and of Science Foundation Ireland under grant numbers 17$/$FRL$/$4832 and SFI$/$12$/$RC$/$2289$\_$P2.}
\thanks{Qiang Wang, David Cordova Bulens, and Stephen J. Redmond are with School of Electrical and Electronic Engineering, University College Dublin, Ireland (e-mail: qiang.wang@ucdconnect.ie; \{david.cordovabulens, stephen.redmond\}@ucd.ie).} 
\thanks{Robert McCarthy is with CeADAR - Ireland’s Centre for Applied AI, University College Dublin, Ireland (e-mail: robert.mccarthy.22@ucl.ac.uk).}
\thanks{Francisco Roldan Sanchez, Kevin McGuinness and Noel E. O’Connor are with School of Electronic Engineering, Dublin City University, Ireland (e-mail: francisco.sanchez@insight-centre.org; \{kevin.mcguinness, noel.oconnor\}@dcu.ie).}
\thanks{\textit{Corresponding to: Stephen J. Redmond}.}
}

\title{\LARGE \bf \vspace{-15pt}
Identifying Expert Behavior in Offline Training Datasets Improves Behavioral Cloning of Robotic Manipulation Policies}

\begin{document}

\maketitle
\thispagestyle{empty}
\pagestyle{empty}
\begin{abstract}
This paper presents our solution for the Real Robot Challenge (RRC) III\footnote{Featured in the NeurIPS 2022 Competition Track, more details see \url{https://real-robot-challenge.com/}}, aiming to address dexterous robotic manipulation tasks through learning from pre-collected offline data. In this competition, participants were given two types of datasets for each task with mixed skill levels: expert and mixed. Each expert dataset is collected by a domain-specific high-skill policy, whereas the mixed dataset is collected using both expert and non-expert policies. During our early exploration of the problem, we found that the simplest offline policy learning algorithm, known as behavioural cloning (BC), exhibited remarkable learning capabilities and learned a very proficient policy with minimal human intervention when trained on expert datasets. Notably, BC outperformed even the most advanced offline reinforcement learning (RL) algorithms. However, when applied to mixed datasets, the performance of BC deteriorates; similarly, the performance of offline RL algorithms is also less than satisfactory. 

Upon examining the provided datasets, it was apparent that each mixed dataset contained a significant proportion of expert data, which should theoretically enable the training of a proficient BC agent. However, the expert data is not labelled in the datasets. As a result, we attempted to filter the expert data from the mixed datasets. Unfortunately, basic intuitive reward-based extraction methods were inadequate for this purpose. Therefore, we propose a classifier trained using semi-supervised learning, and designed to identify the underlying pattern of the expert behaviour within a mixed dataset. The trained classifier can then be utilized to effectively isolate the expert data.

To further boost the BC performance, we take advantage of the geometric symmetry of the RRC arena to augment the training dataset through mathematical transformations. Ultimately, our submission outperformed that of all other participants, even those who employed intricate offline RL algorithms alongside complex data processing and feature engineering.
\end{abstract}

\section{Introduction} \label{section-intro}
Data-driven learning methods are emerging as a promising approach for dexterous robotic manipulation tasks. These methods are capable of learning skillful manipulation control from scratch and have begun to surpass conventional control methods in certain scenarios\cite{rw:sc1,rw:sc2,rw:sc3, dextrous}. However, their application in the physical world remains limited. This is primarily due to the fact that these methods often require extensive data acquisition from the environment, which is typically costly and time-consuming in physical settings. These issues can be mitigated by making use of available pre-collected data. The Real Robot Challenge (RRC) III sought to encourage the development of offline policy learning algorithms that can make efficient use of such pre-existing real-world data \cite{rrc-iclr}, and thus improve the performance of these learning methods when deployed in practical real-world scenarios.

\begin{figure}[t]
\begin{center}
    \begin{minipage}{\columnwidth}
    \centerline{
        \subfigure[TriFinger robot platform]{\label{fig:trifinger-robot}
        \includegraphics[width=0.95\columnwidth]{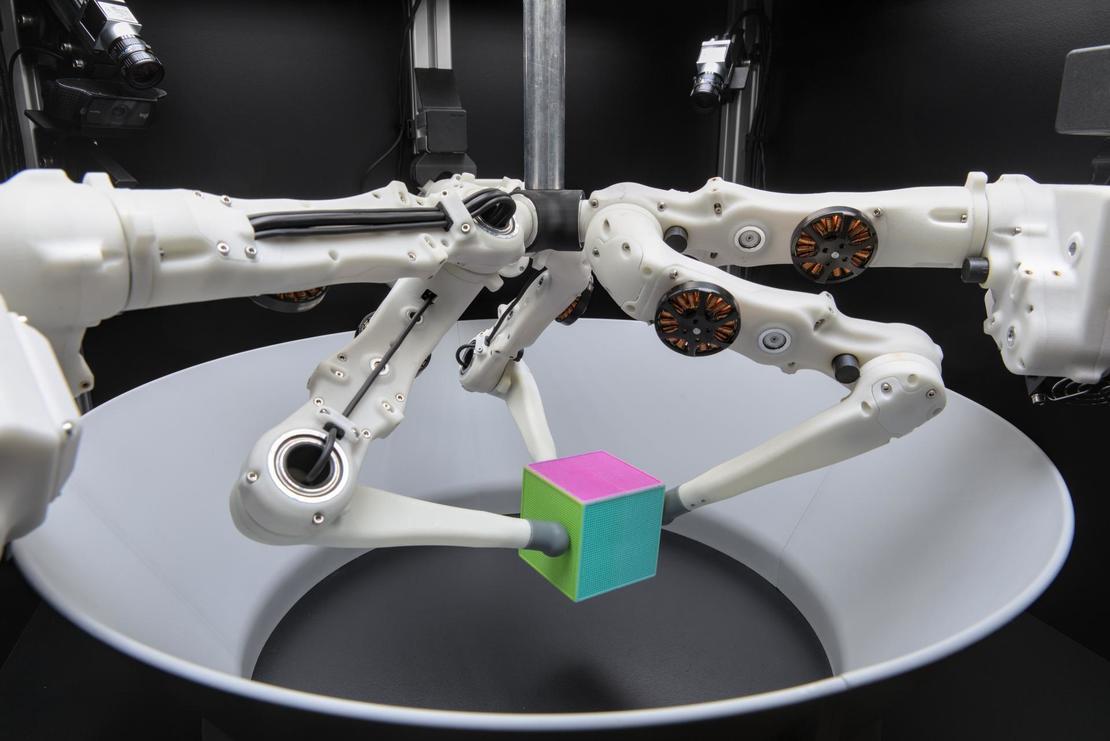}}
        }
    \centerline{
        \subfigure[\textit{Push} task]{\label{fig:robot-push}
        \includegraphics[width=0.46\columnwidth]{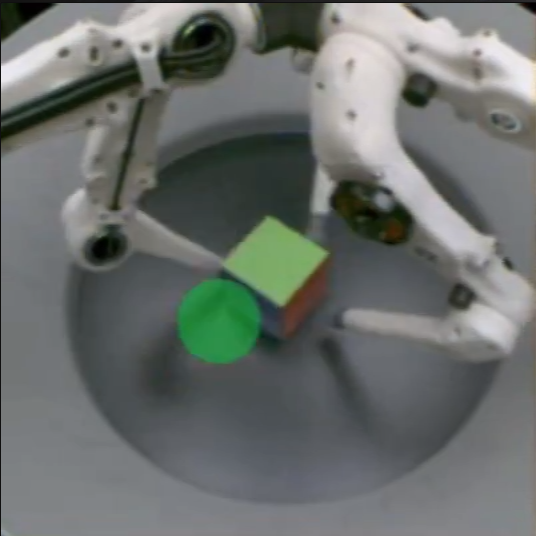}
        }
        \subfigure[\textit{Lift} task]{\label{fig:robot-lift}
        \includegraphics[width=0.46\columnwidth]{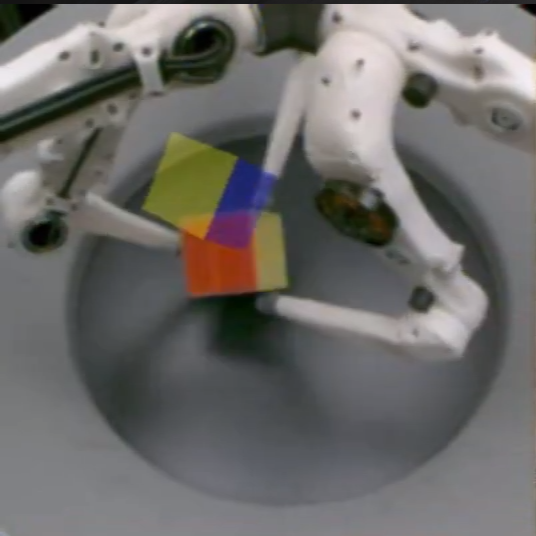}
        }
    }
    \end{minipage}
    \caption{(a) Illustration of the physical TriFinger robot from the RRC III competition. (b) Illustration of the \textit{push} task, where the translucent green dot indicates the 2D target position on the arena floor. (c) Illustration of the \textit{lift} task, where the translucent cube indicates the target 3D position and orientation, often above the arena floor.}
    \label{fig:trifinger-robot-task}
\end{center}
\vspace{-1.5em}
\end{figure}

\subsection{Real Robot Challenge III}
The RRC III robotic platform, as seen in Fig.~\ref{fig:trifinger-robot}, features three identical robotic fingers symmetrically positioned at $120\degree$ intervals around a circular arena. The coloured cube is the object to be moved. In this competition, participants aim to solve two tasks: the \textit{push} task and the \textit{lift} task (see Fig.~\ref{fig:robot-push}-\ref{fig:robot-lift}). The objective of the \textit{push} task is to relocate the cube to a target 2D position on the arena floor. The \textit{lift} task, on the other hand, presents a more demanding challenge, necessitating the cube to be elevated and maintain a stable target pose (3D position and orientation). The reward is determined using a logistic kernel, $k(x)=(b+2)(exp(a\left \| x \right \| ))+b+exp(-a\left \| x \right \| )^{-1}$, which operates on the distance between the desired position and the achieved position (for the \textit{push} task), or the average distance between the desired ehgit corner points and the achieved eight corner points of the cube (for the \textit{lift} task). The parameters $a$ regulate the length scale over which the reward diminishes, while $b$ controls the sensitivity of the reward to small distances. The initial position and orientation of the cube are randomly generated to introduce variability in the starting conditions.

For each task, we are provided two datasets: one obtained from an expert policy (denoted as expert dataset), and the other collected from multiple policies exhibiting varying skill levels (denoted as mixed dataset). As a result, we have a total of four distinct datasets. The specific configurations of each dataset are outlined in TABLE~\ref{table:datasets-config}. The competition rules strictly permit only learning-based approaches and prohibit the merging of the given datasets. Participants are provided with a cluster of six real TriFinger robots and a simulation environment with the same configuration to evaluate their trained policies. It is crucial to emphasize that, in the spirit of offline learning, any data collected during the evaluation, whether from simulation or the real robot, cannot be utilized for refining the policy further. For more detailed information, please refer to the RRC III website provided above.

\begin{table}[h]
\caption{Information about the datasets provided.}
\begin{tabular}{l||ccc}
Dataset       & Mean return & Number of transitions & Episodic length \\ \hline\hline
\textit{Push}/expert & \makebox[4ex][r]{660}         & $2.8\times10^{6}$    & \makebox[4ex][r]{750}             \\
\textit{Push}/mixed  & \makebox[4ex][r]{429}         & $2.8\times10^{6}$    & \makebox[4ex][r]{750}             \\
\textit{Lift}/expert  & \makebox[4ex][r]{1064}        & $3.6\times10^{6}$    & \makebox[4ex][r]{1500}            \\
\textit{Lift}/mixed  & \makebox[4ex][r]{851}         & $3.6\times10^{6}$    & \makebox[4ex][r]{1500}            \\ 
\end{tabular}
\label{table:datasets-config}
\end{table}

\subsection{Dataset investigations}
We explored the provided datasets during the early stage of the competition. We employed cutting-edge offline policy learning algorithms to train multiple agents, specifically including PLAS\cite{plas}, TD3+BC\cite{td3bc} and BC\cite{bc}. The evaluated results are presented in TABLE~\ref{table:comparitive-results}. We found that agents trained on the expert dataset consistently outperformed those trained on the mixed datasets. Notably, BC demonstrated superior performance on the expert datasets, surpassing all sophisticated offline RL algorithms. Furthermore, when training with mixed datasets, none of the algorithms were able to achieve satisfactory performance, especially for the relatively challenging \textit{lift} task. Therefore, we believed that better utilizing the mixed datasets would be key to outperforming other competitors.

\begin{figure}[t]
\begin{center}
    \centerline{
        \subfigure[\textit{Push}/expert and \textit{Push}/mixed]{\label{fig:push-reward}
        \includegraphics[width=0.495\columnwidth]{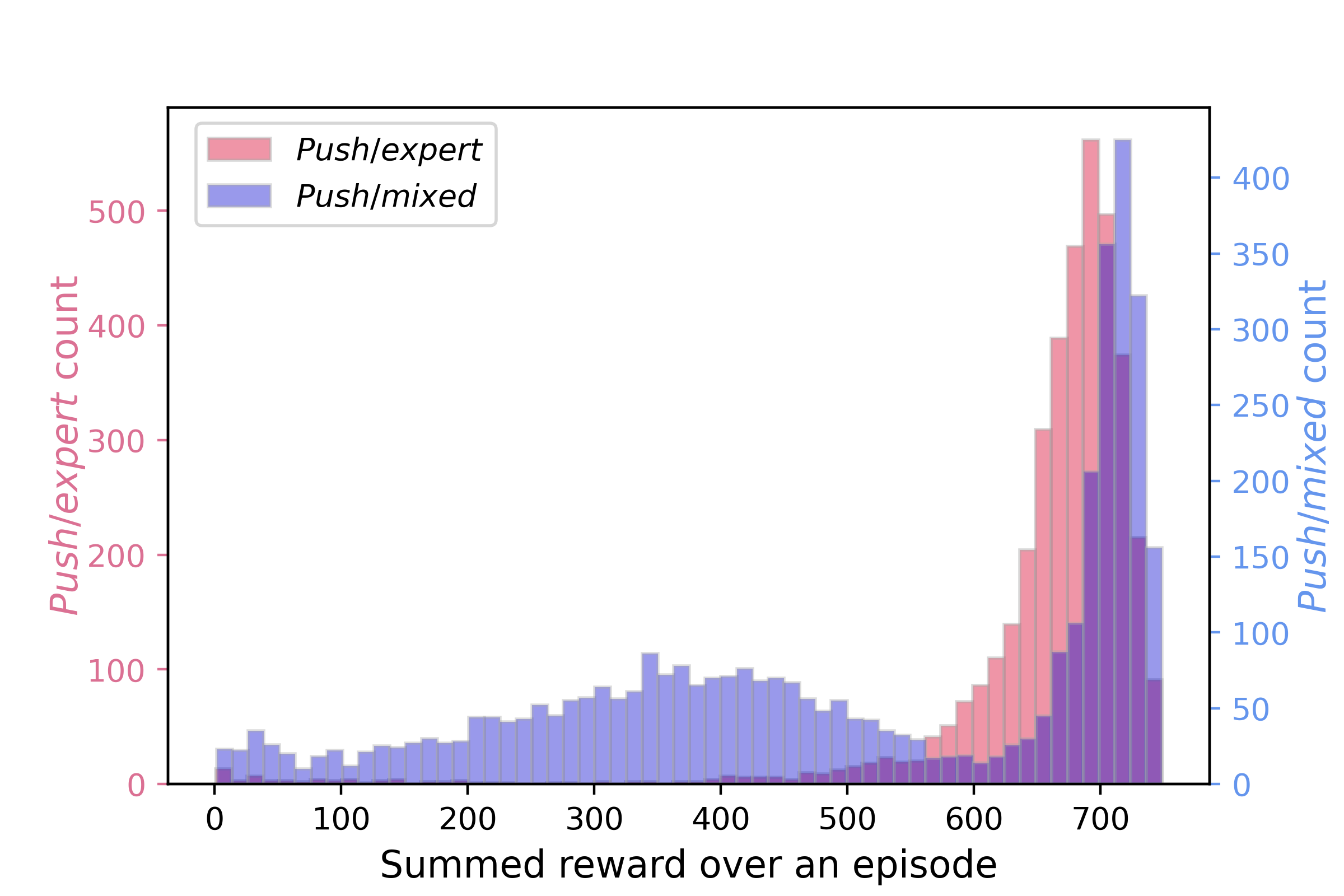}
        }
        \hspace{-10pt}
        \subfigure[\textit{Lift}/expert and \textit{Lift}/mixed]{\label{fig:lift-reward}
        \includegraphics[width=0.495\columnwidth]{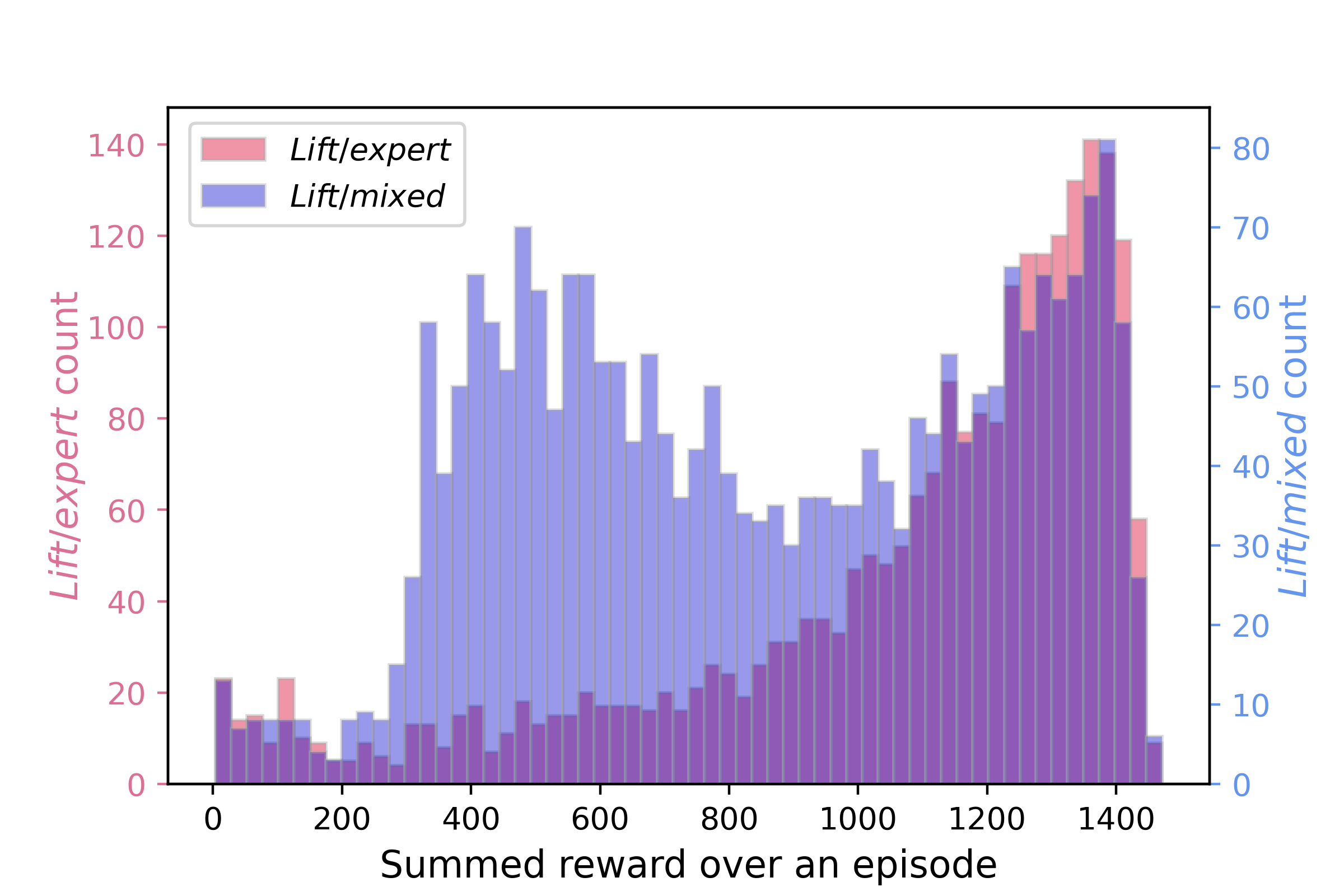}
        }
    }
    \caption{The histograms of the accumulated rewards for all four datasets, calculated  by summing the rewards across all time steps in each episode (each manipulation trajectory). The expert dataset consists mostly of successful episodes achieving large cumulative reward values and it has one distinct peak (depicted by red bars in the two subfigures). The mixed dataset appears to consist of two distinct peaks (depicted by blue bars in the two subfigures).}
    \label{fig:reward-distribution}
\end{center}
\vspace{-1.8em}
\end{figure}

To gain insight into the composition of the mixed datasets, we plot their reward distributions and compare them with those of the expert datasets in Fig.~\ref{fig:reward-distribution}. It is evident that a significant portion of episodes in the mixed datasets achieve expert-level scores. Hence, we formulated a hypothesis that the mixed dataset contains a large proportion of expert data. Motivated by the remarkable performance of BC on the expert dataset, we set out to filter out a subset of expert data from the mixed dataset and subsequently utilize this subset for BC training.

A straightforward method to distinguish expert data within a mixed dataset is to utilize the available rewards. This is because experts are more likely to gain higher cumulative reward when performing a task compared to less effective policies. Hence, in our investigations, we extracted the top 10\% and 50\% most rewarded episodes to be used for BC training. The evaluated results are shown in TABLE~\ref{table:comparitive-results}. This method basically addresses the \textit{push}/mixed dataset as there is a substantial performance disparity between the expert and the weaker policy that are assumed to have generated the dataset (see Fig.~\ref{fig:push-reward}).  However, the effectiveness of this filtering approach diminishes when applied to the \textit{lift}/mixed dataset. This is because the expert and weaker policies exhibit relatively similar performance on the \textit{lift}/mixed task, resulting in significant overlap in cumulative rewards (see Fig.~\ref{fig:lift-reward}). Indeed, using the top 10\% episodes based on reward, a high separation threshold ensures that the majority of the extracted data consists of expert data. However, this approach also results in a smaller expert data subset size, which turns out to not be sufficient for effectively training a robust policy.

\subsection{Our method}
In this paper, we propose a learning-based filter method to address the above problem; it is essentially a binary classifier that can learn the patterns of expert behavior and use the acquired knowledge to better filter expert data from the mixed dataset.

Considering the reduction of the amount of data left for policy learning after filtering, we utilize the rotational symmetry of the RRC III robot platform geometry to augment the training data. Essentially, this method triples the dataset size through a mathematical transformation. Moreover, we propose a theory-to-real transfer training method, enabling the trained policy derived from the theoretically augmented dataset to adapt to the actual environment.



\section{Related works}
\subsection{Real-world dexterous robotic manipulation}
Dexterous robotic manipulation refers to enabling robots to control and manipulate objects with human-like precision and strength. Generally, such a system utilizes artificial sensors to gather information about the state of the world. It also incorporates a controller that processes the sensor inputs and generates control signals. These signals are then transmitted to robotic manipulators with the capability to execute delicate movements akin to those performed by humans \cite{rw:manipulator}, such as pick-and-place operations\cite{rw:pick-and-place, rw:pick-and-rotate}, rotation tasks \cite{rw:pick-and-rotate, rw:rotate-task}, and the utilization of tools \cite{rw:tool-use}. 

\textbf{Traditional robotic manipulation:}  This entails the manual coordination of individual servo systems and their integration with the environment. The objective of these approaches is to empower the robot system to carry out predetermined actions in an anticipated state. Several control objectives are known to improve the compliance of traditional manipulation. These objectives include position control\cite{rw:position-control}, force control\cite{qiang-force-paper}, hybrid position/force control\cite{rw:hybrid-control}, impedance control\cite{rw:impedance-control} and visual servoing control\cite{rw:visual-control}. Traditional approaches were utilized in previous years of the RRC and have achieved success\cite{rw:rrc1-winner}. However, these methods often require extensive manual tailoring, which can lead to limitations in terms of generalization and robustness.

\textbf{Data-driven robotic manipulation:} This aims to facilitate robot movements through the collection and analysis of extensive data. By leveraging machine learning technologies, its objective is to enable robots to autonomously learn and refine their manipulation skills. Deep RL\cite{dqn} has emerged as a popular data-driven approach in the past decade, showcasing remarkable achievements in various complex robotic control domains\cite{rw:sc1,rw:sc2,rw:sc3,rw:sc4}. Notably, it was employed in previous RRC solutions\cite{rrc2021}, outperforming traditional approaches and showcasing impressive results.

\subsection{Offline policy learning}
Deep RL mentioned earlier typically requires a large number of interactions with the environment for each training attempt, however learning a satisfactory policy typically demands multiple attempts; hence this is impractical in many real-world scenarios. A promising solution involves leveraging a pre-collected dataset for multiple training attempts, essentially transforming the control policy learning into an offline paradigm. Offline policy learning research techniques primarily fall into two categories: offline imitation learning (IL) and offline reinforcement learning. For insight surverys and benchmarks, refer to \cite{il-review, offline-rl-review} and \cite{d4rl, rlplug, offline-rl-benchmark} respectively.

\textbf{Offline IL}: BC is a straightforward approach to imitation learning (IL) \cite{bc, bc2}. It seeks to discover a policy that replicates the behavior employed to accomplish a specific task. Typically, the desired behavior for cloning is obtained from an expert source, such as a human \cite{huamandemo1, humandemo2}, or a proficient scripted agent \cite{bcq}. BC can be seen as a form of supervised regression, as it learns a policy that maps states from the dataset to the corresponding actions. BC demonstrates high efficiency when trained with high-quality expert data, leading to outstanding performance of the trained agent \cite{humanoid-control}. In comparison to more complex IL methods like GAIL \cite{gail} and inverse RL \cite{inverserl}, BC generally achieves superior performance. But standard BC, as a form of supervised learning, has certain limitations. Firstly, to avoid regression ambiguity, the action conditioned on a state should be drawn from a unimodal distribution, or the target action mode must constitute the majority proportion in the dataset \cite{bc-tutorial}. Secondly, a reasonably large training dataset is necessary to mitigate the covariate shift issue \cite{covariate-shift}, which refers to the compounded error arising from unseen data during deployment. Lastly, BC's performance is typically constrained by the capabilities of the demonstrator \cite{under-bounded}. In recent times, a promising advancement called implicit BC has been introduced to enhance the performance of BC \cite{ibc}. Implicit BC incorporates the use of energy-based models\cite{enengy-based-model} to achieve this improvement.

\textbf{Offline RL}: This is similar to standard online RL paradigms \cite{rl}, the objective is to discover a policy that maximizes the expected sum of discounted rewards. While most off-policy RL algorithms can be applied offline, the lack of exploration leads to a mismatch between the actual training batch and the expected state-action visitation under the current policy. Consequently, during the policy evaluation phase, state-action pairs that are not present in the dataset would be inaccurately estimated; during the policy improvement stage, the policy may learn to overestimate out-of-distribution (OOD) actions \cite{bcq}. The OOD issue often results in subpar performance of policies learned through classical off-policy RL algorithms in pure offline settings. To address this challenge, various methods have been proposed, including policy regularization \cite{iql, bcq, plas, td3bc} and the use of conservative value estimates \cite{cql, uncertainty}. These approaches aim to minimize the discrepancy between the learned policy and the policy employed by the agent responsible for generating the dataset.

Our algorithm draws inspiration from a branch of offline RL literature, known as filter-based BC \cite{crr, awr, bail, awac, fqi-awr}. These algorithms involve estimating advantage functions using available rewards and utilizing the learn function to perform weighted regression for BC, either with hard or soft weights. In contrast to these algorithms, our algorithm is less reliant on rewards and instead employs a classifier to directly identify expert-like behaviors. As a result, our algorithm outperforms the state-of-the-art offline RL algorithms in all dexterous manipulation tasks in the RRC III competition.

\section{Method}
\subsection{Background}
The problem of offline policy learning can be formulated within the context of a Markov decision process (MDP), denoted as $\mathcal{M}$ = ($\mathcal{S}$, $\mathcal{A}$, $\mathcal{R}$, $\mathcal{P}$, $\gamma$) \cite{mdp}. In this formulation, $\mathcal{S}$ represents the state space, $\mathcal{A}$ corresponds to the action space, $\mathcal{R}$ is the reward function, $\mathcal{P}$ represents the dynamics of the environment, and $\gamma$ denotes the discount factor.  In each time step $t$, the agent receives a state $s_{t} \in \mathcal{S}$ and selects an action $a_{t}\in \mathcal{A}$ based on a policy $\pi(a_{t}\mid s_{t})$. After executing the chosen action in the environment, the agent receives a reward $r_{t} = \mathcal{R}(s_t, a_t)$, and the environment transitions to a new state $s_{t+1}$. In the context of a RRC III, the dataset provided for each task can be represented as $\mathcal{D} = {(s_{t},a_{t},r_{t},s_{t+1})_{t=1...i}}$, where $i$ denotes the number of time steps in the episode.

\subsection{Filtering expert data from mixed datasets}\label{our approch}
We propose a learning-based binary classifier to filter expert data from the mixed datasets; the classifier can learn the behavior pattern of the expert by comparing the expert's behaviour with that of a non-expert. The entire process of data filtering is a semi-supervised learning process, where the trained classifier is continuously used to separate more training samples from the mixed dataset for the next training iteration. This continues until the composition of the training samples no longer changes. The algorithm can be decomposed into the following steps:

\subsubsection{Generate training samples}
First, we manually label a small section of data from the mixed dataset, assuming it contains expert demonstrations; and this subset serves as positive training examples for the binary classifier. As previously mentioned, the highest-rewarding trajectories in the mixed dataset are highly likely to have been generated by an expert. Therefore, in this context, we consider the top 10\% most rewarded episodes as the expert data. The determination of this percentage value also involved referencing Fig.~\ref{fig:lift-reward}, which shows a decreasing trajectory amount trend for weaker policies in the mixed dataset as the episodic reward increases. Intuitively, it appears that there are fewer episodes collected from weaker policies in the highest cumulativereward range. In fact, our experiments found that the choice of this percentage value can vary without detriment, as it will converge to a similar result for initial percentage values ranging from 5\% to 15\%.

On the contrary,  it is necessary to have negative samples that contrast with the positive expert samples in order to generate supervised signals to train the classifier. However, using reward alone is not an effective method for directly identifying valid negative samples from the mixed dataset, particularly for the \textit{lift} task. Fig.~\ref{fig:lift-reward} illustrates that a certain portion of the expert data receives lower scores, resembling a weaker policy. This can be attributed to the complexity and randomness of the environment, which can lead to failures even with strong policies. Thus, we propose an alternative approach, in which we combine states and actions from different sources. The generated negative sample set involves: 1) Pairing states from the positive sample subset with randomly generated actions; 2) Pairing randomly generated states with actions from the positive sample subset; and 3) Pairing randomly-generated states with randomly-generated actions. The new state-action pairs are unlikely to resemble the expert's behavior and this mixture of different sources helps introduce novel and diverse examples into the classifier training process. Actually, any state-action pair that deviates from expert behavior and is generated using a different approach than those we proposed could work here.

\subsubsection{Train the classifier}
We feed state-action pairs into the neural network to train the classifier. It is worth noting that within the RRC III setting, there is a considerable dimensional discrepancy between the state and action spaces. Therefore, we initially deploy a fully-connected encoder to condense the state dimension. Subsequently, we concatenate the action with the reduced-dimension state tensor. This combined tensor is then inputted into a fully-connected predictor. The final layer of this network employs a softmax activation function and its output can be interpreted as probability. Our chosen loss function is binary cross-entropy. 

\subsubsection{Employ the classifier}
We feed the state-action pairs of each time step of an episode in the mixed dataset into a classifier to estimate the probability of that state-action pair being collected by the expert. The RRC III robotic data is organized into episodes, each consisting of a sequence of consecutive time steps. To obtain a more comprehensive prediction for each episode, we average the probabilities estimated for each timestep in the episode. This mean probability is referred to as the confidence (\textit{\textbf{conf}}) in the following.

We establish a confidence threshold, $\bm{\theta_{conf}}$, to binarize the \textit{\textbf{conf}} of each episode.  The resulting binary value determines if the corresponding episode should be identified as expert-collected and subsequently used as a positive sample to further train the filter classifier. Once the membership of filtered  positive samples converges, it serves as the training set for BC. We regard $\bm{\theta_{conf}}$ as a tunable hyperparameter and optimize it by observing the evaluated performance of the policy trained using BC. Our experiments indicate that the optimal value of $\bm{\theta_{conf}}$ is 0.95 for the \textit{lift}/mixed dataset and 0.96 for the \textit{push}/mixed dataset.

\subsection{Symmetry-based data augmentation}
As previously discussed, one of the main drawbacks of BC is its susceptibility to the covariate shift issue due to being a supervised learning method; note, this issue could be mitigated by increasing the quantity or diversity of the training data\cite{shift1,shift2}. Hence, we used data augmentation techniques during the RRC III competition to further improve the performance of BC. In prior work, data has been augmented for offline policy learning by editing the state vector to improve the robustness of the learned policy, using techniques such as adding noise, scaling, dimensional dropout, state-switch, state mix-up, and adversarial transformation \cite{s4rl}; however, these approaches have a limited ability to diversify the dataset, as these operations are anchored around the same state-action pair. In our approach, we exploit the spatial (rotational) symmetry of the robot arena to generate new state-action pairs by transforming each component of the state and action. This allows us to introduce additional variations in the dataset and enhance the diversity of the training samples.

\subsubsection{Rotational transformation}
The three fingers of the robot are evenly spaced around the center of the circular arena, with an angle difference between two adjacent fingers of $120\degree$. Since the structure of each finger is theoretically identical, the correctness of the data, including the states of the object and robot, should remain unchanged after rotating clockwise or counterclockwise around the central point of the arena by $120\degree$. 

In effect, we will spatially rotate the entire experiment (robot, arena, and object) around the centre of the arena by integer multiples of $120\degree$ in the world frame, but the indexes of each finger do not move and still reference the same location in the world frame. Since different transformations are required to perform this rotation, depending on whether we are transforming a robot state or a spatial pose of the object, we split the state vector into robot and object state subvectors ($\textbf{s} = [{\textbf{s}^{robot}}^T, {\textbf{s}^{obj}}^T]^T$)\footnote{More information about the observation space, we direct the reader to the RRC III website provided above.}. Simultaneously, we perform a permutation on the robot state subvector $\textbf{s}^{robot}$ and its associated action vector $\textbf{a}$; and we perform spatial rotation on the object state subvector $\textbf{s}^{obj}$:
\begin{equation}
\label{eq_rot_finger}
    \textbf{s}^{robot}_{aug}(\alpha) = \textbf{s}^{robot}((\alpha + k\cdot120\degree) \hspace{1mm} \% \hspace{1mm} 360\degree) 
\end{equation}
\begin{equation}
\label{eq_rot_action}
    \textbf{a}_{aug}(\alpha) = \textbf{a}((\alpha + k\cdot120\degree) \hspace{1mm} \% \hspace{1mm} 360\degree)
\end{equation}

\vspace{-12pt}

\begin{equation}
\label{eq_rot_obj}
    \begin{bmatrix}s^{obj}_{x,aug} \\ s^{obj}_{y,aug} \end{bmatrix} = \begin{bmatrix}\cos(k\cdot120\degree) & -\sin(k\cdot120\degree) \\ \sin(k\cdot120\degree) & \cos(k\cdot120\degree) \end{bmatrix}\cdot\begin{bmatrix}s^{obj}_{x} \\ s^{obj}_{y} \end{bmatrix},
\end{equation}
where $\alpha\in(0\degree,120\degree,240\degree)$ and $k\in(0,1,2)$, $\textbf{s}^{robot}(\alpha)$ and $\textbf{a}(\alpha)$ represents the state and action subvectors for the finger of the robot located at an angle of $\alpha$ degrees, and $s^{obj}_{x}$ and $s^{obj}_{y}$ represent the $x$ and $y$ coordinates of the object. For example, the entire experiment will be spatially rotated $120\degree$ around the $z$ of the world frame when $k=1$ (anticlockwise, looking top-down). The $z$ coordinate of the object remains unchanged. The data after these rotational transformations are concatenated with the original dataset to form a larger augmented dataset.

\subsubsection{Theory-to-real transfer} \label{subsubsec:theory-to-real-transfer}
Similar to the well-known sim-to-real gap problem (which describes how policies learned in ideal simulations often underperform in the real-world due to lack of consideration of variances in physical properties, such as weight, shape, and friction)\cite{sim2real}, trying to leverage spatial symmetry to perform data augmentation can also fall foul of similar assumptions of ideal physical properties. For example, we might assume that all three fingers of the robot are identical in every way, but we know that this is unlikely to be true, and they might vary in ways such as having different frictional properties, the motors generating different torques in response to a given command, or the sensitivity/calibration of the tactile sensors differing across fingers. 

Therefore, we began by training a BC policy on the augmented dataset, utilizing a higher learning rate and longer training length to establish a more general policy. Subsequently, we fine-tuned this policy using the original, non-augmented dataset with a lower learning rate and reduced training length. This process ensures that the final policy deployed aligns more closely with the physical data distribution observed from the real robots.

\subsection{Method summary}
In our final submission, we utilize BC as the control algorithm across all datasets. The training objective of BC is as follows:
\begin{equation}
    \underset{\pi}{min} \underset{(s_{t},a_{t})\sim\mathcal{D}_{expert}}{\mathbb{E}}\left[-\log\space\pi(a_{t} \mid s_{t})\right].
\label{eq:bc}
\end{equation}

For the \textit{lift}/mixed and \textit{push}/mixed datasets, we first employed the data filtering method to separate out the expert data (we assume). Then, we performed augmentation on these expert data. Finally, we trained the BC policy using the method proposed in Sec.\ref{subsubsec:theory-to-real-transfer}. For two expert datasets, the method is basically the same, except that the expert datasets do not need to go through the filtering step. 

\begin{table*}[t]
\vspace{0.5em}
\centering
\caption{The evaluated scores comparing our method to comparative methods, where task-specific scores are given as mean $\pm$ SD. In the evaluation stage, the policy is randomly deployed on one of six available physical robots, and the goal (position and/or orientation of a cube) is randomly generated. The evaluation for each task lasts $15$ episodes. }
\label{table:comparitive-results}
\resizebox{\textwidth}{!}{
\begin{tabular}{
  l||
  >{\collectcell\text}r<{\endcollectcell}
  @{${}\pm {}$}
  >{\collectcell\text}r<{\endcollectcell}
  >{\collectcell\text}r<{\endcollectcell}
  @{${}\pm {}$}
  >{\collectcell\text}r<{\endcollectcell}
  >{\collectcell\text}r<{\endcollectcell}
  @{${}\pm {}$}
  >{\collectcell\text}r<{\endcollectcell}
  >{\collectcell\text}r<{\endcollectcell}
  @{${}\pm {}$}
  >{\collectcell\text}r<{\endcollectcell}
  >{\collectcell\text}r<{\endcollectcell}
  @{${}\pm {}$}
  >{\collectcell\text}r<{\endcollectcell}
  >{\collectcell\text}r<{\endcollectcell}
  @{${}\pm{}$}
  >{\collectcell\text}r<{\endcollectcell}|
  >{\collectcell\text}r<{\endcollectcell}
  @{${}\pm{}$}
  >{\collectcell\text}r<{\endcollectcell}|
  >{\collectcell\text}r<{\endcollectcell}
  @{${}\pm{}$}
  >{\collectcell\text}r<{\endcollectcell}
  >{\collectcell\text}r<{\endcollectcell}
  @{${}\pm{}$}
  >{\collectcell\text}r<{\endcollectcell}|
  >{\collectcell\text}r<{\endcollectcell}
  @{${}\pm{}$}
  >{\collectcell\text}r<{\endcollectcell}
}

{}         & \multicolumn{2}{c}{BC} & \multicolumn{2}{c}{\text{10\% BC}} & \multicolumn{2}{c}{\text{50\% BC}} & \multicolumn{2}{c}{CRR} & \multicolumn{2}{c}{TD3+BC} & \multicolumn{2}{c|}{PLAS}  & \multicolumn{2}{c|}{C-Aug} & \multicolumn{2}{c}{Ablation 1} & \multicolumn{2}{c|}{Ablation 2} & \multicolumn{2}{c}{\textbf{Ours}}\\ \hline\hline
{\textit{Push}/expert} & \text{626} & \text{101} & \text{-} & \text{-} & \text{-} & \text{-} & \text{611} & \text{127} & \text{623} & \text{99} & \text{618}&\text{92}   & \text{619} &\text{74}   & \text{626} &\text{101}   & \text{641} &\text{47}   & \textbf{662} &\textbf{87} \\

{\textit{Push}/mixed} & \text{497} & \text{88} & \text{541} & \text{90} & \text{623} & \text{43} & \text{599} & \text{93} & \text{604} & \text{111} & \text{595}&\text{121}   & \text{532} &\text{81}   & \text{618} &\text{83}   & \text{627} &\text{88}   & \textbf{636} &\textbf{126} \\

{\textit{Lift}/expert} & \text{928} & \text{205} & \text{-} & \text{-} & \text{-} & \text{-} & \text{792} & \text{227} & \text{852} & \text{401} & \text{874}&\text{359}   & \text{861} &\text{187}   & \text{928} &\text{205}   & \text{1077} &\text{199}   & \textbf{1130} &\textbf{193} \\

{\textit{Lift}/mixed} & \text{489} & \text{282} & \text{503} & \text{117} & \text{492} & \text{219} & \text{606} & \text{312} & \text{698} & \text{362} & \text{707}&\text{350}   & \text{495} &\text{214}   & \text{917} &\text{237}   & \text{980} &\text{280}   & \textbf{1038} &\textbf{305} \\ \hline

{Average} & \text{635} & \text{169} & \text{-} & \text{-} & \text{-} & \text{-} & \text{652} & \text{190} & \text{694} & \text{243} & \text{699}&\text{231}   & \text{622} &\text{139}   & \text{772} &\text{157}   & \text{831} &\text{154}   & \textbf{867} &\textbf{178} \\

\end{tabular}
}
\end{table*}

\section{Experiments and results}
This section will showcase the filtering accuracy of our method in diffrentiating the expert data from mixed datasets. Furthermore, we will emphasize the advantages of incorporating our filtering and augmentation techniques in offline policy learning and compare them with other baseline methods. 

\subsection{Classifier filtering accuracy}
\begin{figure}[t]
\begin{center}
    \centerline{
        \subfigure[\textit{Push}/mixed dataset]{
        \includegraphics[width=0.495\columnwidth]{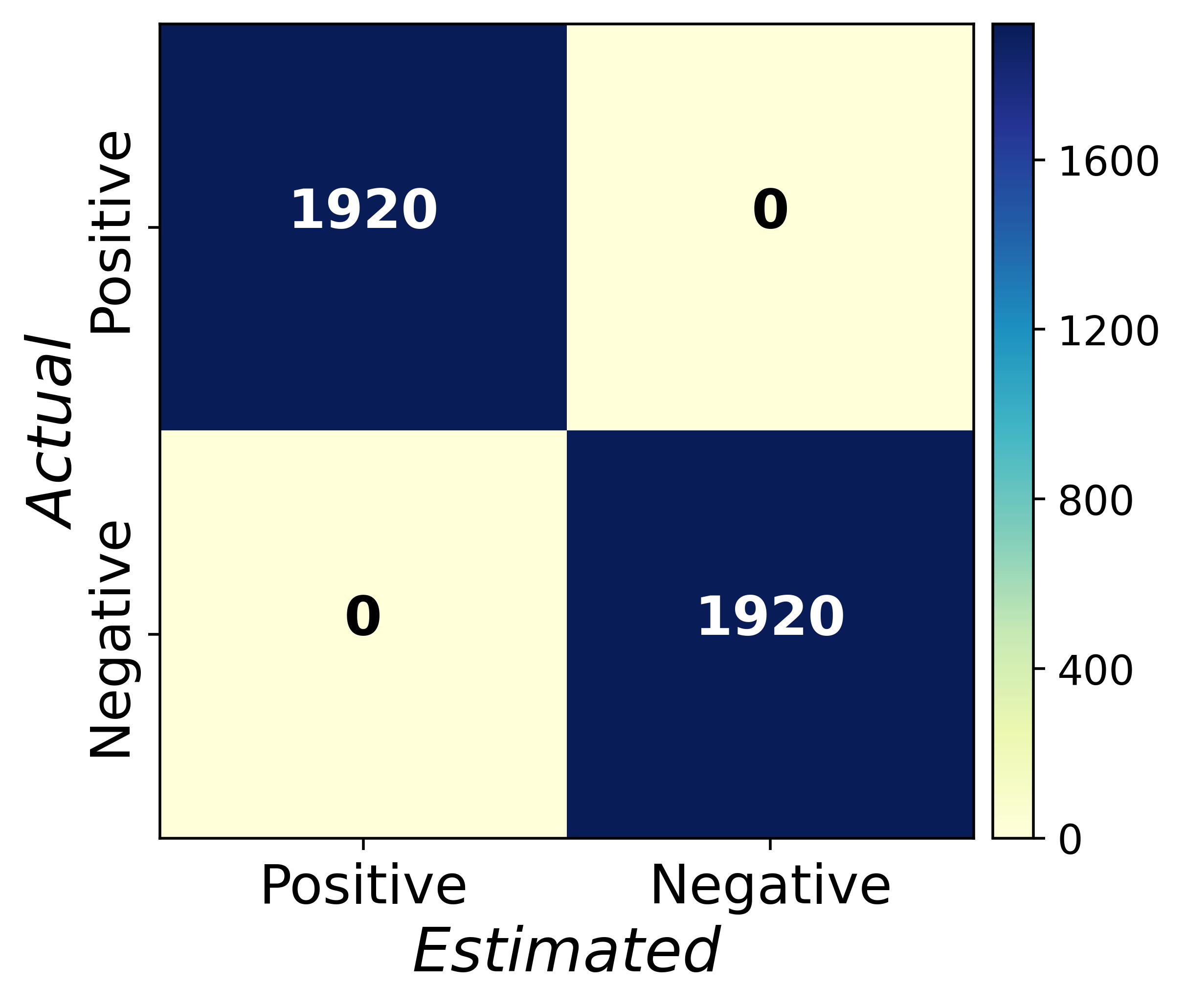}
        }
        \hspace{-10pt}
        \subfigure[\textit{Lift}/mixed dataset]{
        \includegraphics[width=0.495\columnwidth]{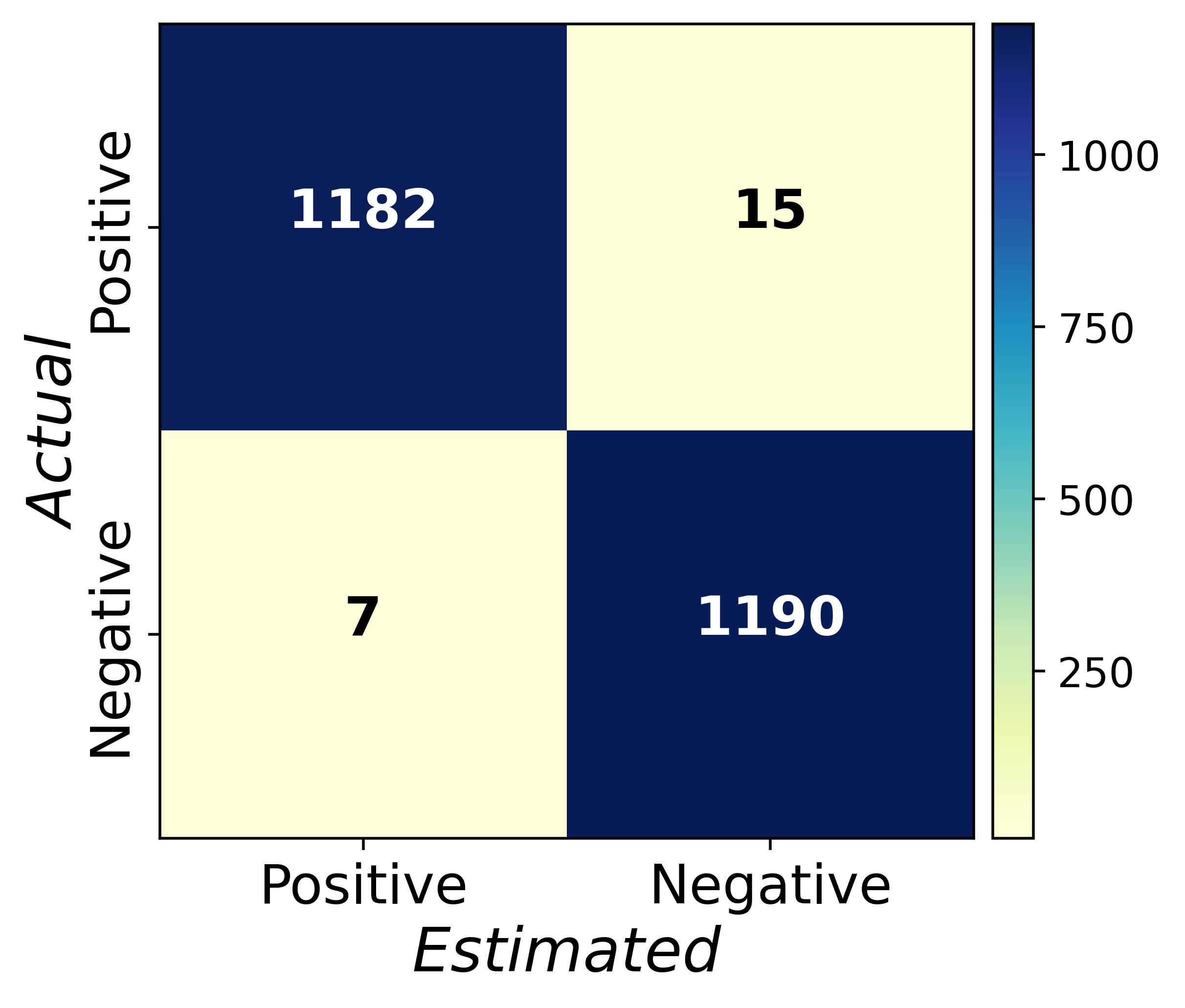}
        }
    }
    \caption{Confusion matrices showing the performance of our filtering method in recognising expert-generated episodes among a larger mixed dataset from the RRC III.}
    \label{fig:confusion-matrixs}
\end{center}
\vspace{-1.5em}
\end{figure}

We reached out to the organizers of RRC III to inquire about the composition of the mixed datasets after the competition. We obtained this information to establish a reliable ground truth for evaluating our classifier. The evaluation results are displayed in Fig.~\ref{fig:confusion-matrixs}. 

Our filtering method achieves exceptional accuracy, even when applied to the \textit{lift}/mixed dataset, where the reward-based performances of expert and weaker policies are notably similar on many episodes. Remarkably, our method achieves 100\% accuracy on the comparatively simpler \textit{push}/mixed dataset. These results substantiate the reliability and effectiveness of our method for practical applications.

\begin{table*}[h]
 \caption{The final official ranking of RRC III competition. The organisers pre-defined several goals in the official evaluation protocol and evenly distributed them among the six robots. Our team is named \textit{excludedrice}. \textit{superiordinosaur} and \textit{jealousjaguar} shared joint third place. The baseline score is the mean episodic return of the dataset. More details refer to the official leaderboard: \url{https://real-robot-challenge.com/leaderboard}}
\begin{tabular}{ll||ccccc}
\multicolumn{1}{l}{} & \multicolumn{1}{l||}{Baseline score} & 660                & 429        & 1064        & 851        & 751     \\ \hline
{\#}                    & \multicolumn{1}{l||}{Team name}     & Push/expert        & Push/mixed & Lift/expert & Lift/mixed & Average \\ \hline\hline
1 & \textbf{excludedrice} & 624 $\pm$ 144 & \textbf{635 $\pm$ 137} & \textbf{956 $\pm$ 431} & \textbf{923 $\pm$ 442} & \textbf{784} \\
2                    & decimalcurlew                & \textbf{639 $\pm$ 112}          & 613 $\pm$ 134  & 841 $\pm$ 415   & 717 $\pm$ 383  & 703   \\
\multirow{2}{*}{3}   & superiordinosaur             & 618 $\pm$ 143          & 575 $\pm$ 191  & 856 $\pm$ 452   & 571 $\pm$ 346  & 655   \\
                     & jealousjaguar                & \textbf{639 $\pm$ 121} & 561 $\pm$ 178  & 855 $\pm$ 392   & 506 $\pm$ 348  & 640   \\ 
\end{tabular}
 \label{table:ranking}
 \centering
\end{table*}

\subsection{Comparative performance evaluation}
\subsubsection{Baseline algorithms}
We firstly compare the performance of our approach with other relevant offline policy algorithms, including:

\begin{itemize}[leftmargin=*]
    \item \textbf{CRR}\cite{crr}: CRR learns the Q-function to construct the advantage function:
        \begin{equation}
            \label{eq_rot_obj}
                \hat{A} (s_{t},a_{t})=Q_{\theta}(s_{t},a_{t})-\frac{1}{m} {\textstyle \sum_{j=1}^{m}}Q_{\theta}(s_{t},a^{j}),
        \end{equation}
        where $a^{j}\sim \pi(\cdot\mid s_{t})$, $\pi(\cdot\mid s_{t})$ is the learned policy, and $Q_{\theta}(\cdot,\cdot)$ refers to the learned critic. The advantage of a specific state-action pair relative to the dataset can be obtained through this function and then be used to weight the importance of the training samples for BC. This method shares similarities with our approach, as both aim to partially focus the BC on the more promising transitions in the dataset.

    \item \textbf{PLAS}\cite{plas}:  A variational autoencoder (VAE) is trained on the raw dataset to capture the underlying distribution of actions. In the policy improvement stage, the policy function outputs a latent action, which is subsequently fed into the VAE's decoder to produce an action that aligns with the distribution of actions in the raw dataset. This method is employed to mitigate the generation of OOD actions.

    \item \textbf{TD3+BC}\cite{td3bc}: This method introduces BC into the TD3 algorithm \cite{td3} as the regularization term:
    \begin{equation}
        \pi = \underset{\pi}{argmax}\underset{(s_{t},a_{t})\sim\mathcal{D}}{\mathbb{E}}[\lambda Q(s,\pi(s))-(\pi(s)-a)^{2}],
    \end{equation}
    It can guide the policy to prioritize actions present in the dataset, effectively avoiding OOD actions.
\end{itemize}

\subsubsection{Baseline augmentation method}
We secondly compare our augmentation method with the classical approach presented in \cite{s4rl}, which involves the addition of Gaussian noise to the state: $s_{aug}=s+\epsilon$, where $\epsilon \sim  \mathcal{N}(0,3\times 10^{-4})$. This technique can simulate real-world variations and uncertainties; as a result, it should enhance the adaptability and robustness of the policy trained by BC, in theory.

\subsubsection{Ablated algorithms}
We finally conduct an ablation study that dissects the contribution of each component to the overall performance of BC, including:
\begin{itemize}[leftmargin=*]
    \item \textbf{Ablation 1}: Only the data from the raw dataset is utilized for training the BC model here, while augmented datasets are not used. Specifically, for mixed datasets, we employ filtered subsets for BC model training without augmentation. For expert datasets, we directly use the raw datasets for BC training without augmentation as well.
    \item \textbf{Ablation 2}: BC is trained exclusively using the augmented dataset without subsequent fine-tuning. Specifically, for the mixed datasets, we augment the filtered subsets for BC model training. For the expert datasets, we directly perform augmentation on the raw datasets for BC model training.  Both experiments are conducted without additionally fine-tuning on the raw data.
\end{itemize}

\subsubsection{Results and analysis}
We present the outcomes of the baselines, ablation studies and our method in TABLE~\ref{table:comparitive-results}. It is evident that CRR, PLAS, and TD3+BC exhibit suboptimal performance on RRC III tasks, particularly on the demanding \textit{lift}/mixed dataset. We propose a hypothesis that offline learning alone lacks the essential capability to accurately approximate the action-value function required for the RRC III environment and tasks due to its complexity.

Traditional data augmentation techniques failed to enhance the performance of BC in the context of RRC III settings (see results of C-aug in TABLE~\ref{table:comparitive-results}). Unexpectedly, it even detracts the overall performance of BC. This decline may be attributable to the inherent complexity of the relationship between state and action. Hence, the introduction of noise via augmentation appears to overtax the neural network's learning process instead of facilitating it.

From the two sets of ablation experiments conducted, it is also evident that our data augmentation approach effectively enhances the performance of BC in addressing the four manipulation tasks. Moreover, additional fine-tuning of BC using raw data subsequently yields further improvement in the model's efficacy. Importantly, these optimization and fine-tuning strategies not only progressively improve performance, but also manage to do so without unduly increasing the model's complexity or computational burden during training; And it do not incur any additional cost during deployment, i.e., the policy is still fast.

\section{Discussion}
Overall, our methods significantly enhance the performance of the vanilla BC algorithm, allowing it to achieve expert-level performance on both mixed datasets. In contrast, complex offline RL algorithms struggle to learn an effective policy, particularly when faced with the high-complexity \textit{lift}/mixed dataset\footnote{A demo video can be found at \url{https://youtu.be/N2xkaGtZtio}}. TABLE~\ref{table:ranking} presents the final official evaluated scores of the RRC III competition, where our team achieved the notable accomplishment of surpassing the baseline in the \textit{lift}/mixed task. The control policy employed by team \textit{decimalcurlew} is TD3+BC \cite{td3bc}. They utilized a method known as spatial smoothing \cite{smooth} to handle noisy data sourced from the physical environment. Similarly, team \textit{superiordinosaur} adopted a BC-based policy learning approach, incorporating feature selection to eliminate superfluous features. Team \textit{jealousjaguar} implemented IQL \cite{iql} for their control policy, augmenting their data with conventional techniques. Their reports are accessible on the leaderboard page of the RRC III website that we provided above.

Our approach exhibits promising potential for generalization. In practical/real-world applications, the datasets often consist of mixed-quality data, characterized by multi-modality or noise. For such situations, it is beneficial to selectively utilize the highest-quality expert data within the mixed-quality dataset (provided that a sufficient amount of expert data is available). This is because offline policy learning algorithms tend to perform better when trained on expert data, rather than on mixed data. In our future work, we aim to broaden our approach to accommodate more complex datasets. Indeed, our filtering method comes with a set of limitations, including cases where it is unable to obtain the initial positive data subset. Furthermore, the labor-intensive process of tuning the confidence threshold $\bm{\theta_{conf}}$ in our technique was noted, prompting us to explore the possibility of using an adaptive $\bm{\theta_{conf}}$ in our future research. Since our approach hinges on supervised classification, it presents us with the opportunity to further improve its effectiveness using strategies such as ensemble techniques.

For the augmentation method, in the physical world, the robotic system and the environment in which it is deployed may occasionally have spatial symmetries; as a simple example, bimanual robots have two arms and usually a left-right mirror symmetry. Combined with our theory-to-real transfer training method we proposed, such spatial symmetries may be useful in finding more general policies which can later be fine-tuned on physically-consistent real-world data.

\section{Conclusion}
To summarize, this paper comprehensively describes and evaluates our solution for the RRC III. We propose an effective strategy for recognizing behaviors produced by a specific expert policy. This technique enables a learning algorithm to exclude behaviors associated with lower-skilled agents, thereby enhancing the effectiveness of the policy learned. In addition, we introduce a geometric data augmentation method capable of significantly boosting the performance of the policy. When paired with our theory-to-real transfer technique, the training outcomes align more accurately with real-world scenarios, with impressive results.



\section*{APPENDIX}
In order to learn the neural network parameters of our filtering method, we employ the Adam optimizer \cite{adam}. We set the learning rate at $10^{-3}$ and the batch size at 1024. Each training iteration is structured to encompass 20 epochs. For the \textit{push}/mixed dataset, training extended over three iterations, while for the \textit{lift}/mixed dataset, it persisted for four iterations.

When training on the augmented dataset for BC, we employed the Adam optimizer with a learning rate of $10^{-3}$. The training process consisted of $5\times10^{5}$ steps, and each step used a batch size of $1024$. For fine-tuning on the raw dataset, we adjusted the training parameters. The process lasted $2\times10^{5}$ steps, with a batch size of $1024$ and a learning rate of $2\times10^{-4}$.

The implementation code for our filtering method and BC can be found on our GitHub page: \url{https://github.com/wq13552463699/Real-Robot-Challenge-2022.git}.

Our implementations of the policy learning algorithms from other compared groups are based on the open-source library d3rlpy\cite{d3rlpy}. We utilized the recommended hyperparameters for our experiments.

Our experiments ran on a PC with an Intel I9-12900F CPU (2.40 GHz $\times$ 16, 32 GB RAM) and an NVIDIA 3090 GPU. 

\section*{ACKNOWLEDGMENT}
We would like to acknowledge the invaluable hardware, software, and technical support provided by the Max Planck Institute for Intelligent Systems in Tübingen/Stuttgart, Germany; and we are deeply grateful for their exceptional organization of RRC III.

\bibliographystyle{ieeetr}
\bibliography{ref}

\end{document}